\documentclass{article}
\usepackage{textcomp}
\usepackage{PRIMEarxiv}
\usepackage{booktabs}
\usepackage[utf8]{inputenc} 
\usepackage[T1]{fontenc}    
\usepackage{hyperref}       
\usepackage{url}            
\usepackage{booktabs}       
\usepackage{amsfonts}       
\usepackage{nicefrac}       
\usepackage{microtype}      
\usepackage{lipsum}
\usepackage{fancyhdr}       
\usepackage{graphicx}       
\graphicspath{{media/}}     
\usepackage{caption}
\title{Ear-Keeper: A Cross-Platform AI System for Rapid and Accurate 
Ear Disease Diagnosis}

\author{
  Feiyan Lu \\
  The Second Affiliated Hospital of Shenzhen University \\
\texttt{lufeiyan86@126.com} \\
   \And
    Yubiao Yue \\
   Xi'an Jiaotong University \\
  \texttt{jiche2020@126.com} \\
   \And
    Zhenzhang Li \\
 Guangdong Polytechnic Normal University \\
  \texttt{zhenzhangli@gpnu.edu.cn} \\
   \And
    Meiping Zhang \\
  Fujian Normal University \\
  \texttt{mpjason@fjnu.edu.cn} \\
   \And
    Fan Zhang \\
  Foshan Sanshui District People's Hospital \\
  \texttt{zhangfanyisheng\_01@163.com} \\
   \And
    Wen Luo \\
  Kangmeihuada (KMHD) GeneTech Co., Ltd \\
  \texttt{luokmhd@foxmail.com} \\
   \And
    Tong Liu \\
  The Second Affiliated Hospital of Shenzhen University \\
  \texttt{byca0217@163.com} \\
     \And
    Jingyong Shi \\
    Zhongshan Gangkou Hospital \\
  \texttt{15915465572@163.com} \\
       \And
    Guang Wang \\
    Yichang Central People's Hospital \\
  \texttt{2388133531@qq.com} \\
     \And
    Xinyu Zeng* \\
    The Second Affiliated Hospital of Shenzhen University \\
  \texttt{2432968@qq.com}
}

\begin{document}
\maketitle

\begin{abstract}
Early and accurate detection systems for ear diseases, powered by deep learning, are essential for preventing hearing impairment and improving population health. However, the limited diversity of existing otoendoscopy datasets and the poor balance between diagnostic accuracy, computational efficiency, and model size have hindered the translation of artificial intelligence (AI) algorithms into healthcare applications. In this study, we constructed a large-scale, multi-center otoendoscopy dataset covering eight common ear diseases and healthy cases. Building upon this resource, we developed Best-EarNet, an ultrafast and lightweight deep learning architecture integrating a novel Local–Global Spatial Feature Fusion Module with a multi-scale supervision strategy, enabling real-time and accurate classification of ear conditions. Leveraging transfer learning, Best-EarNet, with a model size of only 2.94 MB, achieved diagnostic accuracies of 95.23\% on an internal test set (22,581 images) and 92.14\% on an external test set (1,652 images), while requiring only 0.0125 seconds (80 frames per second) to process a single image on a standard CPU. Further subgroup analysis by gender and age showed consistently excellent performance of Best-EarNet across all demographic groups. To enhance clinical interpretability and user trust, we incorporated Grad-CAM-based visualization, highlighting the specific abnormal ear regions contributing to AI predictions. Most importantly, we developed Ear-Keeper, a cross-platform intelligent diagnosis system built upon Best-EarNet, deployable on smartphones, tablets, and personal computers. Ear-Keeper enables public users and healthcare providers to perform comprehensive real-time video–based ear canal screening, supporting early detection and timely intervention of ear diseases.
\end{abstract}


\section{Introduction}
The ear, being one of the most vital organs in the human body, not only governs our auditory perception but also serves the essential function of maintaining bodily equilibrium\cite{1}. Meanwhile, the ear is also an organ that is frequently affected by diseases, with over 500 million people worldwide being affected by ear infections each year\cite{2}. If not detected and treated promptly, ear diseases can lead to hearing loss and lifelong complications\cite{3}. They can also give rise to severe complications such as acute mastoiditis, labyrinthitis, and meningitis, which can have a significant impact on one's quality of life throughout their lifetime\cite{4,5}. In addition, as the most common ear disease, otitis media affects individuals of all age groups and is a leading cause of healthcare visits worldwide. It is responsible for approximately 20,000 deaths each year due to related complications, with the highest mortality rate occurring among children under the age of five\cite{6}. Furthermore, over 80\% of patients experience at least one episode of acute otitis media before the age of 3, and 40\% of patients have six or more recurrent episodes by the age of 7\cite{7}. This often results in many children suffering from hearing loss, which can have long-term negative effects on their learning abilities and future personal development\cite{8}.

As with other diseases, timely detection and appropriate treatment of ear diseases can help prevent the worsening of the condition and avoid unnecessary side effects of certain medications, such as broad-spectrum antibiotics\cite{9,10,11}. In contrast to other disorders, ear diseases examination is a simple, non-invasive process. In clinical practice, otoscopy is a frequently employed ear examination method . However, due to variations in expertise and experience levels, misdiagnosis by healthcare professionals often occurs, especially in non-specialty clinics (pediatrics, general practice, emergency departments, primary care physicians, and telemedicine) and in areas with limited medical resources\cite{12}. Experts can nevertheless misdiagnose conditions to variable degrees, even in specialized otolaryngology departments, with the diagnosis accuracy for subtypes of middle ear illnesses, for instance, being approximately 75\%\cite{13}. At the same time, the demand from patients will continue to rise, but there won't be enough ear specialists to handle it. For those with ear issues, this lack of ear doctors may make it difficult to provide prompt diagnosis and adequate treatment\cite{2,14}. Especially for children in developing countries, substantial investments are needed to provide facilities for ear disease detection and treatment to prevent long-term hearing impairments and other sequelae \cite{15}. Accessible and affordable ear healthcare services are crucial to ensure early diagnosis, appropriate treatment, and prevention of complications related to ear diseases in children.

To alleviate the aforementioned challenges, there is an urgent need to develop a new generation of diagnosis devices for ear diseases that can provide timely, convenient, and accurate diagnoses. Fortunately, the ear endoscope, commonly used in ear examinations, is a readily replaceable device that can be easily operated without specialized training. This presents a feasibility for integrating self-recognition capabilities into the device using emerging deep learning techniques. While some research has already utilized deep learning models for intelligent recognition of various types of ear endoscopic images, achieving promising recognition results, these efforts often lack extensive data validation and only focus on model development and accuracy, without considering the applicability of the models in real-world scenarios. In reality, an intelligent identification model for ear illnesses should balance the trade-offs among diagnosis accuracy, model parameter size, and model response speed in addition to focusing on diagnosis performance. Repeated observations of the afflicted ear canal areas are frequently necessary for clinical diagnosis. As a result, if a model can achieve extremely quick inference speed, it can let both individual users and otolaryngologists perform thorough and in-depth scans of ear lesions in real-time video format. This is more rigorous and scientific than depending only on intelligent recognition based on a single static image, and it is in line with the actual clinical diagnosis process. Moreover, such a model can be deployed on edge devices with limited computational capabilities, such as low-end computers, smartphones, and Pad. Small model parameter sizes imply minimal memory and storage requirements, enabling easy deployment on resource-constrained devices like smartphones and embedded systems. Specifically, the key to using artificial intelligence for diagnosing ear diseases lies in developing applications that possess the aforementioned features and are compatible with different devices. Such applications do not require expensive computing devices or network services and can be reliably applied to various local devices and scenarios, aiding in the early detection and treatment of ear diseases for a wide range of patients. This, in turn, effectively reduces the overall impact and losses caused by ear diseases.

In this work, we aim to design a model that can be easily used in various real-world scenarios and develop a corresponding intelligent diagnosis system for ear lesions. To achieve this, we first establish the largest dataset of ear lesion images ever created. We then employ the well-known lightweight network ShuffleNetV2\_X0\_5 \cite{16} as our base model and propose a novel image feature fusion module called the Local-Global Spatial Feature Fusion Module (LGSFF). By combining ShuffleNetV2\_X0\_5 with a multi-auxiliary loss accumulation strategy and LGSFF, we design a network named Best-EarNet that achieves a good trade-off between diagnosis performance, inference speed, and model parameter size. Best-EarNet demonstrates super-fast inference speed and extremely small model parameter size, making it easily deployable in different scenarios. Importantly, Best-EarNet achieves excellent diagnosis performance for eight types of ear diseases as well as normal ears, validated through diverse populations of different genders, age groups, and clinical settings. In the final stage of our work, we develop an application called Ear-Keeper. Ear-Keeper consists of four versions: a mobile version based on edge computing, a mobile version based on cloud servers, a Pad version based on edge computing, and a PC version based on edge computing. The mobile versions primarily assist general users in self-diagnosis and regular check-ups. With the aid of a compact electronic otoscope, users can promptly diagnose their ears when experiencing discomfort. Simultaneously, the Pad and PC versions can assist healthcare professionals in diagnosing patients in various settings, including community screenings, community clinics, resource-limited areas, and specialized otolaryngology practices. This helps reduce misdiagnosis, improve diagnosis efficiency, enhance the patient experience, and reduce the substantial manpower and material costs associated with traditional diagnosis. Additionally, we employ the Grad-CAM method to visualize the decision-making process of the Best-EarNet model. This allows users and medical practitioners to better understand the decision-making process of the deep learning model and helps developers validate, debug, and improve the model's behavior more effectively.

\section{Related Work}
In the research of intelligent recognition of ear diseases, intelligent recognition has been performed based on ear endoscope images, focusing on features such as the eardrum and ear canal. Here, we summarize some prominent related studies. In terms of methods, the main approaches include machine learning and deep learning methods. In machine learning, neural network (NN), decision tree (DT), support vector machine (SVM), and K-nearest neighbor (K-NN) have been employed.
In 2017, Senaras et al. conducted a binary classification study on 247 ear endoscope images using an NN model, achieving an accuracy of 84.6\%. In 2018, Myburgh et al. employed a DT+NN method to classify 389 ear endoscope images into five categories, achieving an accuracy of 81.6\% and suggesting the potential application on smartphone devices. In 2019, Livingstone et al. used an AutoML model to classify 1366 ear endoscope images into eight categories, achieving an accuracy of 88.7\%. In 2020, Viscaino et al. employed an SVM+K-NN+DT model to classify 1060 ear endoscope images into four categories, achieving an accuracy of 93.9\% and suggesting the potential application on PC devices.

In the domain of deep learning methods, Cha et al. (2019) utilized the Inception-V3+ResNet101 model ensemble to conduct a 6-class classification study on 10,544 ear endoscope images, achieving an accuracy of 93.7\%. Khan et al. (2020) employed the ResNet50 model to classify 2,484 ear endoscope images into three categories, achieving an accuracy of 95\%. Zafer et al. (2020) utilized the VGG16 model to classify 857 ear endoscope images into four categories, achieving an accuracy of 99.5\%. Wu et al. (2021) employed the Xception+MobileNetV2 model ensemble approach to conduct a binary classification study on 12,230 ear endoscope images, achieving an accuracy of 90.7\% and suggesting the potential application on smartphone devices. Zeng et al. (2021) utilized the DenseNet161+169 model ensemble approach to classify 20,542 ear endoscope images into eight categories, achieving an accuracy of 95.6\% and suggesting the potential application on PC devices. Chen et al. (2022) used the MobileNetV2 model to conduct a 10-class classification study on 2,171 ear endoscope images, achieving an accuracy of 97.6\% and suggesting the potential application on smartphone devices. K. Manju et al. (2022) employed the VGG19 model to classify 1,880 ear endoscope images into three categories, achieving an accuracy of 97\%. Habib et al. (2023) used the Vision transformer model to conduct a binary classification study on 1,842 ear endoscope images, achieving an accuracy of 92\%.
\begin{table}[h]
\caption{Previous studies on ear diseases.}
\label{tab:table1} 
\resizebox{\textwidth}{!}{
\begin{tabular}{@{}cccccccc@{}}
\toprule
\textbf{Reference} & \textbf{Number} & \textbf{Category} & \textbf{Accuracy} & \textbf{Year} & \textbf{Model}        & \textbf{Apply} & \textbf{Cross Validation} \\ \midrule
{Senaras\cite{17}}           & 247             & 2                 & 84.60\%           & 2017          & NN                    & ×              & ×                         \\
{Myburgh\cite{18}}           & 389             & 5                 & 81.60\%           & 2018          & DT+NN                 & Smartphone     & ×                         \\
{Cha\cite{19}}           & 10,544          & 6                 & 93.70\%           & 2019          & InceptionV3+ResNet101 & ×              & 5-Fold                    \\
{Livingstone\cite{20}}           & 1,366           & 8                 & 88.70\%           & 2019          & AutoML                & ×              & ×                         \\
{Khan\cite{21}]}           & 2,484           & 3                 & 95.00\%           & 2020          & ResNet50              & ×              & ×                         \\
{Zafer\cite{22}}           & 857             & 4                 & 99.50\%           & 2020          & VGG16                 & ×              & 10-Fold                   \\
{Viscaino\cite{23}}           & 1,060           & 4                 & 93.90\%           & 2020          & SVM, K-NN, DT         & PC             & ×                         \\
{Wu\cite{24}}           & 12,230          & 2                 & 90.70\%           & 2021          & Xception+MobileNetV2  & Smartphone     & ×                         \\
{Zeng\cite{25}}           & 20,542          & 8                 & 95.60\%           & 2021          & DenseNet161+169       & PC             & ×                         \\
{Chen\cite{26}}           & 2,171           & 10                & 97.60\%           & 2022          & MobileNetV2           & Smartphone     & ×                         \\
{Manju\cite{27}}           & 1,880           & 3                 & 97.00\%           & 2022          & VGG19                 & ×              & ×                         \\
{Habib\cite{28}}           & 1,842           & 2                 & 92.00\%           & 2023          & Vision Transformer    & ×              & 5-Fold                    \\ \bottomrule
\end{tabular}}
\end{table}

We summarized the progress and limitations of previous work in Table 1. Overall, deep learning models have achieved higher accuracy compared to traditional machine learning methods. However, direct comparison based solely on accuracy is not entirely appropriate due to variations in the number of classes and images across studies. Therefore, from the perspective of practical applications, we have identified the following four limitations based on the summary of previous work. (1) The limited quantity and variety of datasets have resulted in inadequate generalization of the models, rendering them unsuitable for real-world scenarios. (2) The consideration of model parameters and inference speed has been lacking, leading to the use of models that are not easily deployable on resource-constrained devices such as smartphones and Pad. For instance, models like VGG19 and Vision Transformer pose challenges for deployment on such devices. (3) Ear diseases are clearly influenced by factors such as age and gender, yet most studies have not validated their models on diverse populations or conducted external validations. (4) While some studies have considered deploying the models on specific application devices such as smartphones and PCs, there is a lack of corresponding demonstrations and research on application performance.

\section{Materials and methods}
\subsection{Data Source}
The data used in this study consists of two parts. The first part, called the SZH Dataset, comprises otoscope images from 65,475 patients with ear diseases who visited the Otolaryngology Department of the Second Affiliated Hospital of Shenzhen University from July 1, 2016, to December 31, 2022 (Table 2, 3). The second part, named the FSH Dataset, includes otoscope images from 1,652 patients with ear diseases who visited the Otolaryngology Department of the People's Hospital of Sanshui District, Foshan City, from June 1, 2022, to January 31, 2023 (Table 2). The research protocol of this study was approved by the Review Committee of the Second Affiliated Hospital of Shenzhen University (Approval No: BY-EC-SOP-006-01.0-A01) and the Review Committee of Foshan Sanshui District People's Hospital (Approval No: SRY-KY-2023045). The Helsinki Declaration's tenets were scrupulously followed throughout the study to respect the rights, privacy, and anonymity of the subjects. The use of de-identified data precluded the use of informed consent.
\begin{table}[]
\centering
\caption{Distribution of ear lesion images. SZH represents Shenzhen Bao'an District People's Hospital and FSH represents Foshan Sanshui District People's Hospital.}
\begin{tabular}{@{}ccccccccccc@{}}
\toprule
               & \textbf{AOM} & \textbf{CME} & \textbf{CSOM} & \textbf{EACB} & \textbf{IC} & \textbf{NE} & \textbf{OE} & \textbf{SOM} & \textbf{TMC} & \textbf{Total} \\ \midrule
\textbf{SZH}   & 439          & 548          & 4021          & 451           & 6058        & 4685        & 2507        & 2720         & 1152         & 22581          \\
\textbf{FSH}   & 12           & 17           & 114           & 105           & 427         & 714         & 97          & 91           & 75           & 1652           \\
\textbf{Total} & 451          & 565          & 4135          & 556           & 6485        & 5399        & 2604        & 2811         & 1227         & 24233          \\ \bottomrule
\end{tabular}
\end{table}

\begin{table}[]
\centering
\caption{The distribution of SZH Dataset by gender (M: Male, F: Female) and age.}
\begin{tabular}{@{}ccccccccccc@{}}
\toprule
\textbf{Classes (M)56.05\%} & \textbf{(0,10{]}} & \textbf{(10,20{]}} & \textbf{(20,30{]}} & \textbf{(30,40{]}} & \textbf{(40,50{]}} & \textbf{(50,60{]}} & \textbf{(60,70{]}} & \textbf{(70,80{]}} & \textbf{\textgreater{}80} & \textbf{Total} \\ \midrule
CME                         & 4                 & 47                 & 129                & 58                 & 35                 & 4                  & 23                 & 4                  & 4                         & 308            \\
CSOM                        & 3                 & 139                & 677                & 575                & 447                & 217                & 128                & 18                 & 50                        & 2254           \\
EACB                        & 39                & 13                 & 52                 & 64                 & 29                 & 26                 & 4                  & 13                 & 13                        & 253            \\
IC                          & 1194              & 318                & 1179               & 305                & 216                & 76                 & 64                 & 25                 & 13                        & 3390           \\
OE                          & 8                 & 64                 & 618                & 380                & 218                & 75                 & 27                 & 10                 & 5                         & 1405           \\
SOM                         & 432               & 70                 & 177                & 286                & 242                & 202                & 89                 & 23                 & 2                         & 1523           \\
TMC                         & 5                 & 5                  & 175                & 180                & 129                & 81                 & 48                 & 16                 & 5                         & 644            \\
AOM                         & 60                & 42                 & 20                 & 25                 & 33                 & 20                 & 27                 & 13                 & 5                         & 245            \\
NE                          & 132               & 181                & 947                & 660                & 377                & 211                & 86                 & 24                 & 6                         & 2624           \\\bottomrule
\rule{0pt}{2ex} 
\textbf{Classes (F)43.95\%} & \textbf{(0,10{]}} & \textbf{(10,20{]}} & \textbf{(20,30{]}} & \textbf{(30,40{]}} & \textbf{(40,50{]}} & \textbf{(50,60{]}} & \textbf{(60,70{]}} & \textbf{(70,80{]}} & \textbf{\textgreater{}80} & \textbf{Total} \\ 
CME                         & 10                & 20                 & 41                 & 71                 & 41                 & 31                 & 20                 & 3                  & 3                         & 240            \\
CSOM                        & 2                 & 63                 & 489                & 442                & 343                & 251                & 121                & 36                 & 20                        & 1767           \\
EACB                        & 19                & 3                  & 77                 & 38                 & 19                 & 10                 & 19                 & 10                 & 3                         & 198            \\
IC                          & 527               & 589                & 822                & 186                & 248                & 171                & 62                 & 47                 & 16                        & 2668           \\
OE                          & 10                & 85                 & 434                & 258                & 158                & 99                 & 35                 & 16                 & 7                         & 1102           \\
SOM                         & 397               & 43                 & 151                & 189                & 140                & 137                & 114                & 23                 & 3                         & 1197           \\
TMC                         & 12                & 12                 & 60                 & 167                & 96                 & 120                & 36                 & 4                  & 1                         & 508            \\
AOM                         & 39                & 41                 & 20                 & 17                 & 31                 & 14                 & 18                 & 10                 & 4                         & 194            \\
NE                          & 92                & 139                & 712                & 465                & 293                & 218                & 105                & 28                 & 9                         & 2061           \\ \bottomrule
\end{tabular}
\end{table}

All images in this study were classified into nine categories by four ear specialists with more than 15 years of clinical experience: Acute otitis media (AOM, Fig. 1a), Cholesteatoma of middle ear (CME, Fig. 1b), Chronic suppurative otitis media (CSOM, Fig. 1c), External auditory canal bleeding (EACB, Fig. 1d), Impacted cerumen (IC, Fig. 1e), Normal eardrum (NE, Fig. 1f), Otomycosis external (OE, Fig. 1g), Secretory otitis media (SOM, Fig. 1h), and Tympanic membrane calcification (TMC, Fig. 1i). To ensure diverse data sources, each patient's data was selected to a maximum of two ear endoscope images. During the screening process, the expert panel, removed any ambiguous or indistinguishable images.
\begin{figure}[ht]
    \centering
    \includegraphics[width=0.8\linewidth]{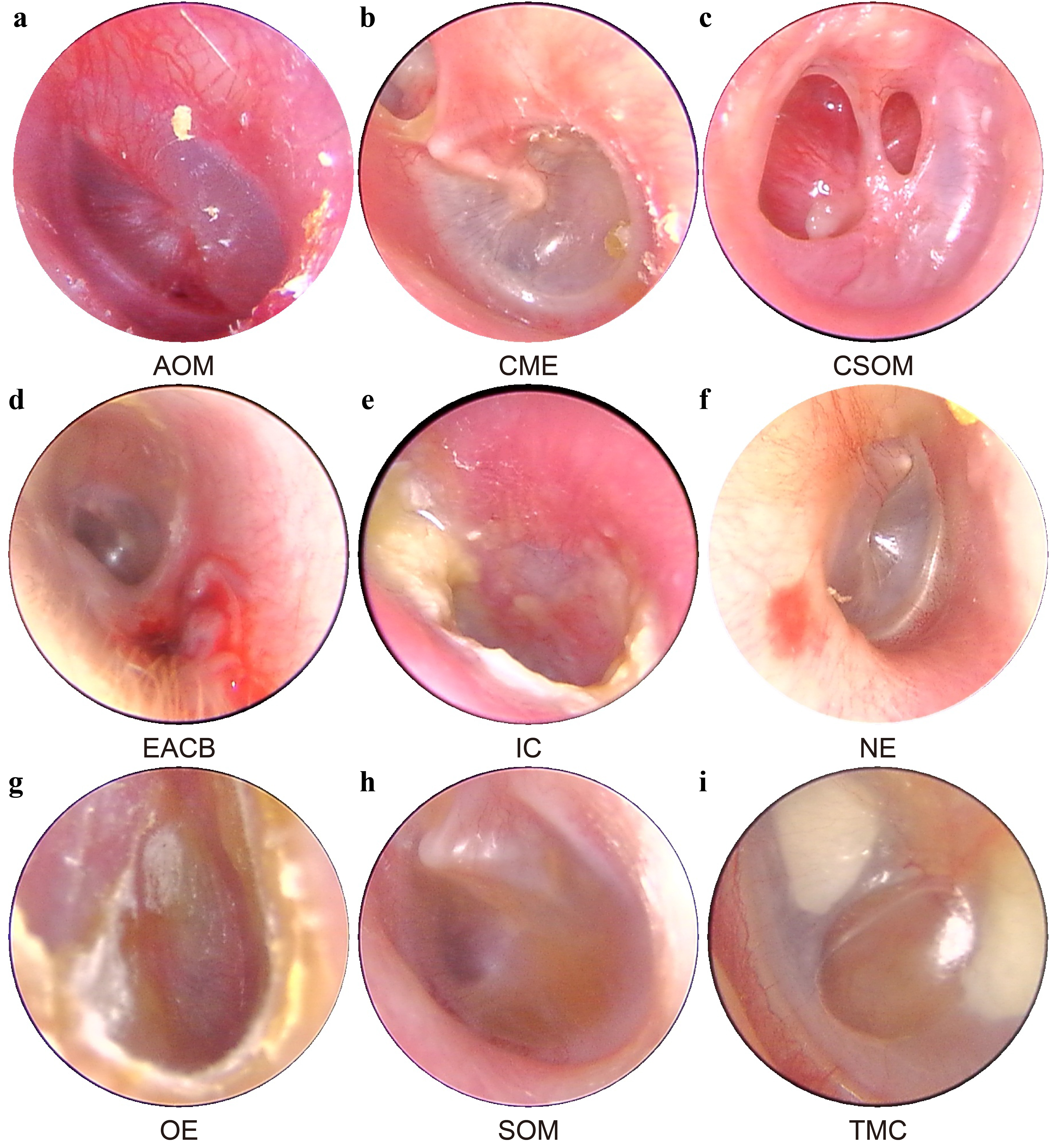}
    \caption{The samples of eight types of ear diseases and normal ear. (a) Acute otitis media  (AOM); (b) Cholesteatoma of middle ear (CME); (c) Chronic suppurative otitis media (CSOM); (d) External auditory canal bleeding (EACB); (e) Impacted cerumen (IC); (f) Normal eardrum (NE); (g) Otomycosis external (OE); (h) Secretory otitis media (SOM); (i) Tympanic membrane calcification (TMC).}
    \label{fig1}
\end{figure}

The SZH Dataset was used for five-fold cross-validation, ensuring the reliability and performance evaluation of the developed model. On the other hand, the FSH Dataset was specifically reserved for testing the model's generalization capability and real-world application performance, simulating its performance in new and unseen data. This separation of datasets allows for a comprehensive assessment of the model's effectiveness.

\subsection{Best-EarNet Design}
In the practical application of the model, the inference speed of the model is particularly important. If the speed is fast enough, the model can assist doctors in the real-time inference of the captured otoscope video. Compared to performing inference on single static images, real-time video inference is more scientifically rigorous. Furthermore, in the actual diagnosis process, doctors often need to use the otoscope to observe the entire human ear canal to ensure a thorough examination. Therefore, the real-time inference is more aligned with the practical clinical diagnosis scenario.
In this study, we conducted a preliminary evaluation of the average frames per second (FPS) and model parameter size of 16 deep learning models. We selected the well-known lightweight network ShuffleNet\_V2\_X0\_5 as the baseline for this work. Then, we improved it using the LGSFF (Local-Global Spatial Feature Fusion Module) and multiple auxiliary loss accumulation strategies to strike a balance between diagnosis performance, average FPS, and model parameters size. The improved network was named Best-EarNet. In the following sections, the authors will comprehensively introduce the design concept and details of Best-EarNet.
\subsubsection{The overall framework of Best-EarNet}
Fig. 2 illustrated the overall architecture of Best-EarNet. The internal structures of the Stem and Stage 2 to Stage 4 (Fig. 2a-b) components are consistent with ShuffleNetV2\_X0\_5. This enables the network to easily load pre-trained weights from the ImageNet dataset during the training process, which helps accelerate model training and improve its performance on specific tasks. In Best-EarNet, each input image (referred to as "input") is first scaled to 3×224×224 dimensions. The image's three channels are then normalized using mean [0.485, 0.456, 0.406] and standard deviation [0.229, 0.224, 0.225].
\begin{figure}[ht]
    \centering
    \includegraphics[width=1\linewidth]{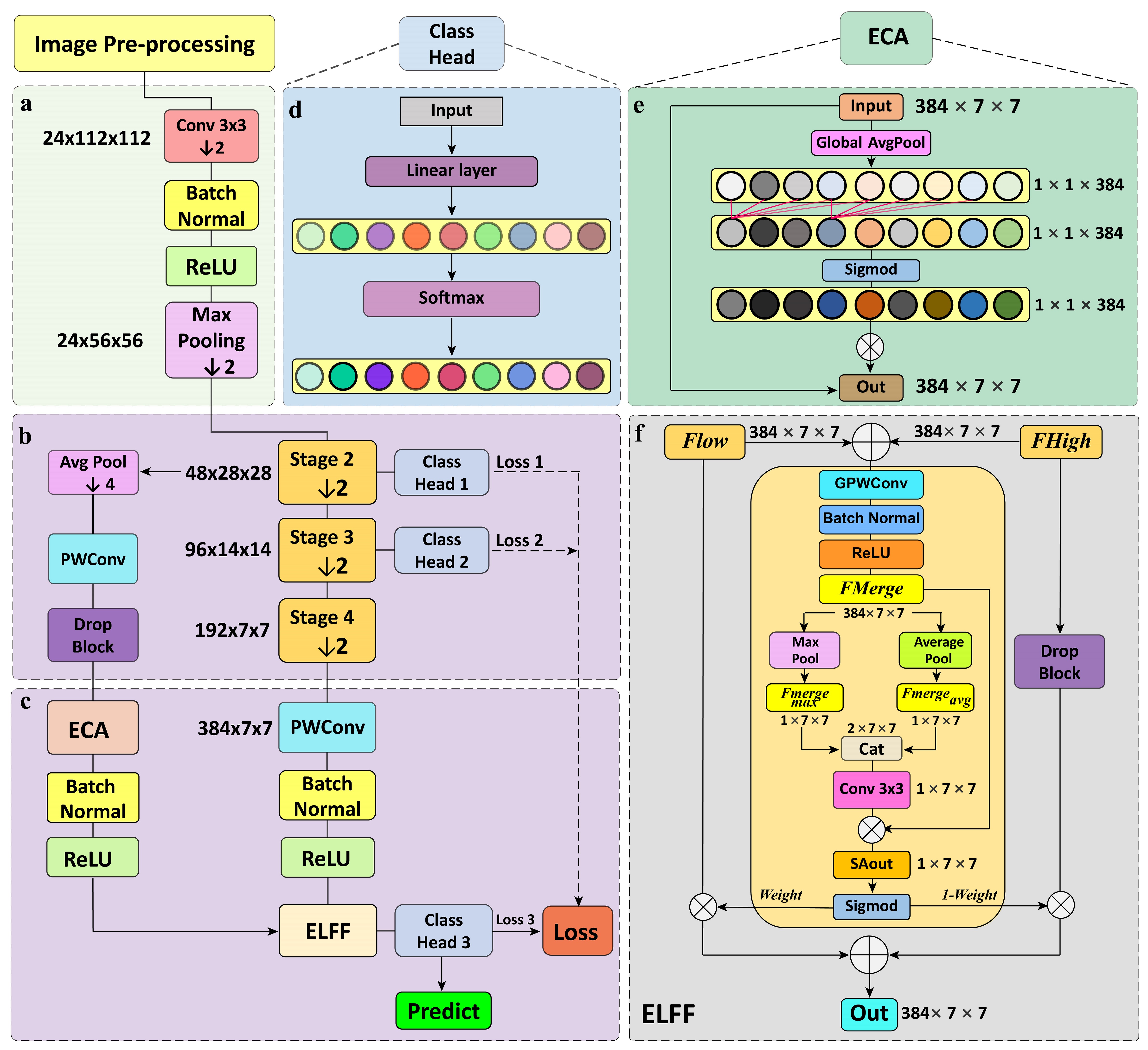}
    \caption{The proposed Best-EarNet architecture. Here, PWconv is Point-wise Convolution, Avg Pool is Average Pooling and GPWconv is Group Point-wise Convolution. (a) The stem of Best-EarNet; (b) The backbone of Best-EarNet and the top half of branch path; (c) The feature fusion stage of Best-EarNet and the bottom half of branch path; (d) the detailed structure of Class Head; (e) the detailed structure of Efficient Channel Attention (ECA); (f) the detailed structure of Local-Global Spatial Feature Fusion Module (LGSFF).}
    \label{fig2}
\end{figure}
Next, the input undergoes 4× downsampling in the stem of Best-EarNet (Fig. 2a), reducing the feature map size from 3×224×224 to 24×56×56. Subsequently, Best-EarNet's Stage 2 to Stage 4 (Fig. 2b) performs downsampling and feature extraction on the input sequentially. After the output of Stage 2, a branch path (Fig. 2b-c) further extracts features from the feature map with richer low-level semantic information, producing an output referred to as "Flow". The size of Flow is transformed from 48×28×28 to 384×7×7 and will be used in the subsequent feature fusion module. At the end of Best-EarNet's backbone, the output of Stage 4 undergoes a final high-level semantic feature extraction using a convolutional layer with a kernel size of 3x3, followed by Batch Normalization and ReLU activation function(Fig. 2c). The output is denoted as "Fhigh". Finally, the LGSFF (Local-Global Spatial Feature Fusion Module) (Fig. 2f) is used to fuse the features from Flow and Fhigh. The fused output is passed to Class Head 3 for result prediction. It is important to note that two auxiliary prediction heads, Class Head 1 and Class Head 2 (Fig. 2d), are added after the outputs of Stage 2 and Stage 3, respectively. They are used to calculate Loss 1 and Loss 2 from low-level semantic information. Loss 1 and Loss 2 are accumulated into Loss 3, aiding in better and faster optimization of the Best-EarNet's model weights.
\subsubsection{Local-Global Spatial Feature Fusion Module}
For ear disease images, the distinctions among various categories are manifested in subtle features such as spatial microstructures, texture patterns, or local color distributions. These fine feature differences might not be evident at the raw pixel level. Furthermore, the inherent similarity in the basic structure and shape of the ear introduces a considerable degree of macroscopic visual homogeneity in these images. As a result, different ear disease images exhibit a high degree of overlap in the feature space, thereby increasing the complexity of the classification task. Additionally, smaller model capacity can lead to insufficient feature extraction capability of the network on images. To address this challenge, we introduced a novel module called the Local-Global Spatial Feature Fusion Module (LGSFF) to fuse low-level and high-level feature information in ear lesion images. LGSFF takes two inputs: Fhigh, which is typically the deep output of the network, and Flow, which is typically the shallow output of the network. Here, we choose to use the output from Stage 2 as Flow and the final output of the network as Fhigh. In our case, both Flow and Fhigh have dimensions of 384×7×7.

LGSFF starts by performing an element-wise addition, GPWconv, BatchNormal and Relu of Flow and Fhigh, resulting in Fmerge. To enhance the model's perception of spatial information in the input image and improve model performance, we use a Spatial Attention Module\cite{29} to process Fmerge and obtain SAout. And then the Sigmod activation function is computed, resulting in activation weights (weight and 1-weight). We then multiply weight with Flow and 1-weight with Fhigh. Before multiplying Fhigh with 1-weight, we apply Dropblock\cite{30}, a regularization technique, to Fhigh. Dropblock randomly drops some blocks, i.e., groups of adjacent feature map units, instead of randomly dropping individual neurons like Dropout. By forcing the model to use other non-dropped regions to learn image features, Dropblock provides stronger regularization, improving the model's robustness and generalization capability. Finally, we add Fhigh and Flow to obtain the output, denoted as Out. The output dimensions of EFLL, i.e., Out, are the same as Flow and Fhigh, which is 384×7×7. The overall process of LGSFF (Local-Global Spatial Feature Fusion Module) can be summarized as Equation (1) to Equation (3).
\[Fmerge = Relu(BN(GPWconv(Flow \oplus Fhigh)))(1)\]
\[SAout = Conv3 \times 3(Cat(Maxpool(Fmerge);Averagepool(Fmerge)))(2)\]
\[Out = (Sigmod(SAout)) \otimes Flow + (1 - (Sigmod(SAout))) \otimes DropBlock(Fhigh)(3)\]
\subsubsection{Processing of branch path and multi-scale supervision}
For Flow, it is not directly passed into LGSFF but processed through a separate branch path. Firstly, an Average Pooling Layer and a Point-wise Convolution Layer are applied to resize the feature map of Flow from 48×28×48 to 384×7×7. Then, DropBlock is used to perform regularization on the feature map. After that, Efficient Channel Attention (ECA)\cite{31} is applied to enable the network to focus more on the low-level feature information in Flow. Batch Normalization and ReLU activation functions are applied to the Flow features. The internal structure of ECA is shown in the diagram, where K=5.

Multi-scale supervision is particularly important for processing complex medical images, as these images may contain structures and details on multiple scales. Therefore, to encourage the network to learn discriminative features at multiple levels, we use multiple auxiliary classification heads to supervise feature representations at different levels, which helps improve the learning efficiency and final performance of the entire network. Concretely, we add auxiliary classification heads to stage-2's ouput and stage-3's ouput of Best-EarNet. Compared to a single classification head, using multiple auxiliary classification heads introduces additional losses from the shallow layers of the network, which increases the gradient signals propagated during backpropagation. This provides extra regularization, improves model robustness, and helps the network converge better and optimize its parameters. The internal structure of these auxiliary heads is shown in the Fig. 2d.
\subsection{Metrics adopted in this work}
n this study, we will assess the model's ability to perform effectively in practical applications by using the following seven metrics.
Accuracy (A): Accuracy is a fundamental evaluation metric that measures the proportion of correctly classified samples by the model, expressed as the ratio of the number of correctly classified samples to the total number of samples. Precision (P): Precision quantifies the proportion of samples predicted as positive by the model that are truly positive. It focuses on the accuracy of positive predictions and is calculated by dividing the number of true positive samples by the total number of samples predicted as positive. Recall (R): Recall (or sensitivity) is defined as the proportion of true positive samples correctly predicted as positive by the model, relative to the total number of true positive samples.  It emphasizes the model's ability to identify as many true positives as possible and is computed by dividing the number of true positive samples by the total number of true positive samples. Specificity (S): Specificity refers to the proportion of truly negative samples that are accurately predicted as negative by the model. It assesses the prediction accuracy of negative samples and is calculated by dividing the number of true negative samples by the total number of true negative samples. F1-score (F1): F1-score combines precision and recall into a single metric that balances both aspects. It is the harmonic mean of precision and recall and serves as a comprehensive performance measure. The aforementioned five metrics can be represented by Equation (4) to Equation (8).
\[A = \frac{{TP + TN}}{{TP + TN + FP + FN}}({\rm{4}})\]
\[P = \frac{{TP}}{{TP + FP}}({\rm{5}})\]
\[R = \frac{{TP}}{{TP + FN}}{\kern 1pt} ({\rm{6}}){\kern 1pt} {\kern 1pt} {\kern 1pt} {\kern 1pt} {\kern 1pt} {\kern 1pt} \]
\[S = \frac{{TN}}{{TN + FP}}({\rm{7}})\]
\[F1 = \frac{{2Precision \times Recall}}{{Precision + Recall}}({\rm{8}})\]
The remaining two metrics are the average Frames Per Second and model parameters size. The average Frames Per Second (Avg FPS) holds significant importance for deep learning models in practical applications. A high Avg FPS implies that the model can process input data at a faster rate, enabling faster real-time prediction or inference. This is crucial for applications that require rapid response, such as real-time video analysis and autonomous driving. Here, a higher average FPS allows users to diagnose their ear lesions via real-time video. The model parameters size is directly associated with the model's size and storage requirements. Larger parameter counts tend to occupy more storage space, which needs to be taken into consideration in resource-constrained environments. In scenarios such as mobile devices and edge computing, the model's size and storage requirements can become limiting factors.

In the process of model evaluation, it is necessary to consider a comprehensive assessment of both the accuracy metrics of the model's computational results and the performance metrics such as average FPS (Frames Per Second). Additionally, it is important to take into account the specific metrics for different diseases, particularly in the case of multi-class models. Therefore, to achieve a more effective model ranking score evaluation, this study proposes a novel approach called the Overall Ranking Score (ORS) for assessing the comprehensive performance of multi-class models. The ORS method is designed to measure the model's performance in practical applications and it is calculated from Equation (9) to Equation (13).
\[OR{S_j} = {\alpha _R}R_j^{RSN} + {\alpha _{F1}}F1_j^{RSN} + {\alpha _P}P_j^{RSN} + {\alpha _S}S_j^{RSN} + {\alpha _A}A_j^{RSN} + {\alpha _{FPS}}FPS_j^{RSN}({\rm{9}})\]
\[f_{ij}^{RS} = \frac{{M_{ij}^f - M_{min\_i}^f}}{{M_{max\_i}^f - M_{min\_i}^f}}\frac{1}{{{L_{IC_{ij}^f}}}},f \in \left( {R,F{\rm{1}},P,S} \right)({\rm{10}})\]
\[f_{i}^{RS} = \sum\limits_{j = 1}^n {f_{ij}^{RS}} ,f_i^{RSN} = \frac{1}{m}\sum\limits_{i = 1}^m {\frac{{f_{ij}^{RS}}}{{f_{i}^{RS}}}} ({\rm{11}})\]
\[A_j^{RS} = \frac{{M_j^A - M_{min}^A}}{{M_{max}^A - M_{min}^A}}\frac{1}{{{L_{IC_j^A}}}},A_j^{RSN} = \frac{{A_j^{RS}}}{{\sum\limits_{j = 1}^n {A_j^{RS}} }}({\rm{12}})\]
\[FPS_j^{^{RSN}} = \frac{{FP{S_j}}}{{\sum\limits_{j = 1}^n {FP{S_j}} }}(13)\]
Here, \(i\) represents the category (1-9), and \(j\) represents the model (1-17). \(M_{ij}^f\) represents the mean value of the multi-fold cross-computation results for metric \(f\), category \(i\), and model \(j\). \(M_{max\_i}^f\), \(M_{min\_i}^f\) represent the maximum and minimum values of metric \(f\) for category \(i\). \(L_{IC_{ji}^f}\) represents the length of the 95\% confidence interval of the multi-fold cross-computation results for metric \(f\), category \(i\), and model \(j\). The superscript \(RSN\) (Ranking Score Normalization) of \(A_j^{RSN}\) indicates the normalization of \(RS\) (Ranking Score). The coefficient \( \alpha \) represents the importance level of each metric, which can be set based on different scenarios. Here, we set it to 1, indicating equal importance.

\section{Results}
\subsection{The workflow of this study}
\begin{figure}[ht]
    \centering
    \includegraphics[width=0.99\linewidth]{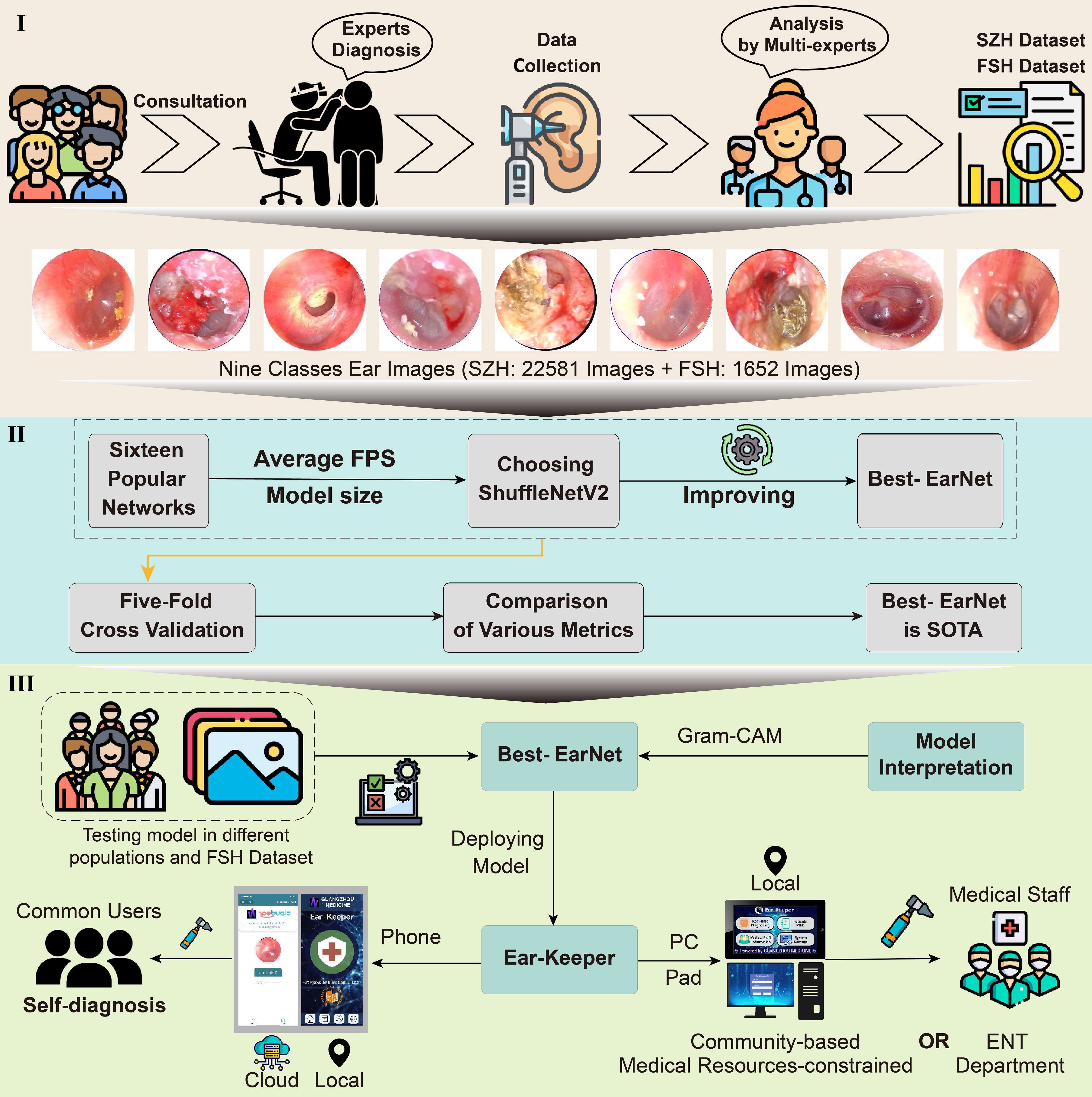}
    \caption{The workflow of this work. (\uppercase\expandafter{\romannumeral1}): Data collection and labeling. (\uppercase\expandafter{\romannumeral2}): Model design. (\uppercase\expandafter{\romannumeral3}): Application development.}
    \label{fig3}
\end{figure}
The workflow of this work as shown in Fig. 3, the entire workflow is divided into three parts. (\uppercase\expandafter{\romannumeral1}) We established a large and diverse dataset of ear lesion images, which consists of 9 categories and a total of 24,233 ear images. The dataset was collected and verified by experts. (\uppercase\expandafter{\romannumeral2}) We aimed to develop a model that can be deployed on any device. Based on previous related work and deployment requirements, we selected 16 networks. Subsequently, considering the model's inference speed and parameter size, we chose ShuffleNet\_V2\_X0\_5 as the baseline model and made improvements to it. The improved model was named Best-EarNet. Through five-fold cross-validation, we found that Best-EarNet is the state-of-the-art model when considering both practicality and diagnosis capability. (\uppercase\expandafter{\romannumeral3}) We evaluated the Best-EarNet model across different genders and age groups based on the clinical information of the images. This ensured the clinical applicability of Best-EarNet, providing reliable support for users in different backgrounds during practical usage and making the model's real-world application more convincing.

Additionally, we further tested the generalization ability of Best-EarNet using an additional test set (FSH Dataset) and applied Grad-CAM\cite{32} to visualize the model's decision-making process. Finally, our team developed four versions of the Ear-Keeper application for different devices: Mobile Ear-Keeper (Cloud), Mobile Ear-Keeper (Local), Pad Ear-Keeper (Local), and PC Ear-Keeper (Local). Due to the outstanding performance of Best-EarNet, the four versions of Ear-Keeper are suitable for different users in various scenarios. Mobile Ear-Keeper (Cloud) and Mobile Ear-Keeper (Local) target general users, allowing individuals to perform self-examinations of their ear canals anytime and anywhere when experiencing ear discomfort. Pad Ear-Keeper and PC Ear-Keeper target doctors and aim to alleviate the current issues in ear disease diagnosis, such as misdiagnosis and an insufficient number of doctors, thus greatly improving diagnosis efficiency and reducing various costs associated with the traditional diagnosis.
\subsection{Experimental setup and model frame selection}
During the entire experimental computation, we fixed the hyperparameters such as learning rate, loss function, optimizer, epoch, and batch size, as shown in Table 4. Moreover, all models were evaluated using five-fold cross-validation. The deep learning framework used in this work was PyTorch, and the experiments were conducted on an Ubuntu 22.04 platform with an Nvidia A40 GPU. The average FPS measurement was performed on a laptop with an Intel (R) Core(TM) i7-10870H CPU @ 2.20GHz. 

To obtain suitable model architectures for our design, we selected networks from the following 16 options. These networks encompass state-of-the-art lightweight models as well as those used in relevant works. These networks are DenseNet161\cite{33}, DenseNet169\cite{33}, FasterNet-T0\cite{34}, InceptionV3\cite{35}, MobileFormer\cite{36}, MobileNetV2\cite{37}, MobileNetV3-small\cite{38}, MobileOne-S0\cite{39}, MobileVit-xxs\cite{40}, ResNet50\cite{41}, ResNet101\cite{41}, ShuffleNetV2\_X0\_5\cite{16}, VGG16\cite{42}, VGG19\cite{42}, Vision in Transformer-Base (ViT-Base)\cite{43} and Xception\cite{44}.

Here, ShuffleNetV2\_X0\_5 was chosen as the baseline with an average FPS of 76 and a parameter count of 1.4 million. To enhance the practicality of the models and ensure fair comparisons, we initialized each network using pretrained weights from the ImageNet dataset. This initialization strategy effectively reduces the training time of the models and alleviates the issue of class imbalance, leading to improved performance and generalization ability.
\begin{table}[]
\centering
\caption{Hyperparameters adopted in this work.}
\begin{tabular}{@{}cc@{}}
\toprule
\textbf{Hyperparameters} & \textbf{Values} \\ \midrule
Optimizer                & Adam            \\
Loss Function            & Cross-entropy   \\
Epoch                    & 100             \\
Batch Size               & 128             \\
Learning Rate            & 0.001           \\ \bottomrule
\end{tabular}
\end{table}

\subsection{Comparison of Models}
We first calculated the Avg FPS and model parameters size of Best-EarNet, which were 0.77M and 80, respectively. Then, under the same experimental conditions, we conducted five-fold cross-validation on ShuffleNetV2\_X0\_5 and Best-EarNet. The confusion matrices of the two networks achieving the best accuracy with and without using ImageNet pretraining in the five-fold cross-validation are shown in Fig.4 (a-d). The results from the confusion matrices indicate that when using pretraining weights, Best-EarNet achieved slightly lower diagnosis accuracy for IC compared to ShuffleNetV2\_X0\_5, but showed significant improvements in diagnosing other classes, particularly AOM and CME, with 8\% and 9\% improvements, respectively.
\begin{figure}[ht]
    \centering
    \includegraphics[width=0.99\linewidth]{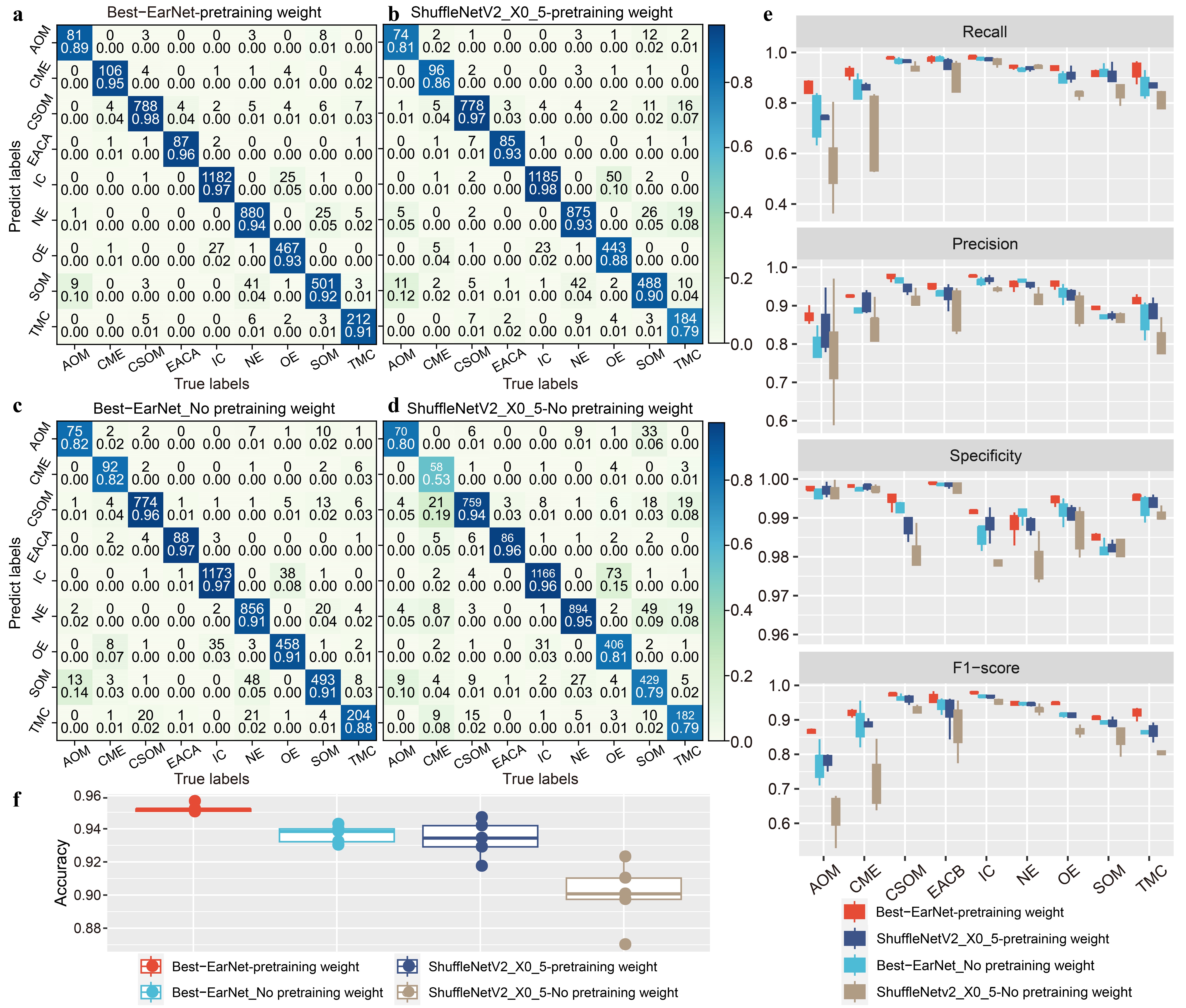}
    \caption{Comparison of various metrics between ShuffleNetV2\_X0\_5 and Best-EarNet. (a-b) The confusion matrix of two networks achieving best Accuracy. (f) The box plots for Accuracy. (e) The box plots of Recall, Precision, Specificity and F1-score.}
    \label{fig4}
\end{figure}
Without using pretraining weights, Best-EarNet had lower diagnosis accuracy for NE compared to ShuffleNetV2\_X0\_5, but showed significant improvements in diagnosing other classes, with CME improving by 29\%. Similar results were observed for the other models from the rest folds(Fig. S1). Additionally, Fig. 4e displays the Avg accuracy box plot for these four models. The results show that Best-EarNet, with or without pretraining weights, achieved higher Avg accuracy (95.23\% and 93.67\%, respectively) compared to the corresponding ShuffleNetV2\_X0\_5 (93.34\% and 90.03\%), demonstrating better performance and stability.

To gain a more detailed understanding of the results for each class, we analyzed the Recall, Precision, Specificity, and F1-score for each class, as shown in the box plot in Fig. 4f. The results indicate that under the five-fold cross-validation, regardless of whether pretraining weights were used or not, Best-EarNet outperformed ShuffleNetV2\_X0\_5 for all nine classes and exhibited lower variability across the five folds. These metric results demonstrate that Best-EarNet exhibits superior diagnosis capability and stability, suggesting the effectiveness of our improvements. It is noteworthy that Best-EarNet achieved nearly a 50\% reduction in parameter size while also experiencing a slight improvement in average FPS.
Continuing from there, we performed five-fold cross-validation on the remaining fifteen networks. Table 5 summarized the performance of Best-EarNet, ShuffleNetV2\_X0\_5, and the other 15 networks across various metrics. Here, each metric represents the average value after five-fold cross-validation. From the analysis of Accuracy, Precision, Recall, Specificity, and F1-score, Best-EarNet achieved commendable results, although it may not have achieved the highest average performance in all metrics. Notably, Best-EarNet surpassed the renowned ResNet50 in terms of Accuracy and outperformed it significantly in terms of average FPS and model parameters size. Furthermore, Best-EarNet exhibited the smallest Mean Standard Deviation (MSD) in Precision, Recall, F1-score, and Specificity. To further evaluate the performance differences among the models, we analyzed the 95\% Confidence Interval (CI) of Recall(Fig. 5), F1-score (Fig. S2), Precision (Fig. S3), Accuracy (Table. S1), and Specificity (Table S2) for all models. The results demonstrated that, similar to MSD, Best-EarNet consistently displayed superior performance in the 95\% CI of each class, indicating its robustness. 

However, in the MSD and 95\% CI results for Accuracy, Precision, Recall, Specificity, and F1-score, some models exhibited similar results, making it challenging to rank and evaluate the models effectively. To provide a comprehensive ranking of the models, we analyzed them based on the proposed ORS and RSN. The findings revealed that Best-EarNet consistently ranked among the top models across all metrics (Fig. 5b), particularly excelling in the ORS metric (Fig. 5c), where it achieved State-of-the-Art (SOTA) performance.
\begin{table}[]
\centering
\caption{Performance of various models on various metrics.}
\resizebox{\textwidth}{!}{
\begin{tabular}{@{}cccccccc@{}}
\toprule
\textbf{Models} & \textbf{\begin{tabular}[c]{@{}c@{}}Avg FPS\\    \\ (CPU)\end{tabular}} & \textbf{Model Params} & \textbf{Accuracy MSD(\%)} & \textbf{Precision MSD(\%)} & \textbf{Recall MSD(\%)} & \textbf{\begin{tabular}[c]{@{}c@{}}Specificity\\    \\ MSD(\%)\end{tabular}} & \textbf{\begin{tabular}[c]{@{}c@{}}F1-score\\    \\ MSD(\%)\end{tabular}} \\ \midrule
Best-EarNet     & 80                                                                     & 0.8M                  & 95.23±0.25                & 93.52±1.26                 & 93.53±1.88              & 99.38±0.12                                                                   & 93.49±0.99                                                                \\
ShuffleNetV2    & 78                                                                     & 1.4M                  & 93.39±1.14                & 91.27±3.00                 & 89.60±3.78              & 99.14±0.29                                                                   & 90.26±2.07                                                                \\
MobileNetV3     & 60                                                                     & 2.5M                  & 95.21±0.46                & 93.87±1.45                 & 93.29±1.87              & 99.37±0.15                                                                   & 93.54±0.97                                                                \\
FasterNet       & 55                                                                     & 3.9M                  & 95.41±0.27                & 94.46±1.71                 & 93.41±2.45              & 99.39±0.16                                                                   & 93.86±1.12                                                                \\
MobileFormer    & 32                                                                     & 3.2M                  & 94.57±0.53                & 93.68±2.36                 & 92.30±2.98              & 99.31±0.20                                                                   & 92.68±1.21                                                                \\
MobileVit\_xxs  & 25                                                                     & 1.3M                  & 95.33±0.36                & 93.33±2.32                 & 94.00±2.29              & 99.42±0.20                                                                   & 93.59±1.37                                                                \\
Xception        & 23                                                                     & 22.0M                 & 95.68±0.34                & 94.63±1.62                 & 94.19±1.92              & 99.43±0.12                                                                   & 92.40±5.75                                                                \\
MobileNetV2     & 22                                                                     & 3.5M                  & 95.13±0.34                & 93.48±2.36                 & 93.64±1.84              & 99.36±0.16                                                                   & 93.50±1.22                                                                \\
InceptionV3     & 16                                                                     & 27.2M                 & 95.13±0.54                & 93.89±2.20                 & 93.22±2.24              & 99.36±0.16                                                                   & 93.56±1.36                                                                \\
MobileOne       & 14                                                                     & 5.3M                  & 95.02±0.17                & 91.85±7.28                 & 92.95±2.95              & 99.35±0.26                                                                   & 93.27±1.50                                                                \\
ResNet50        & 10                                                                     & 25.6M                 & 95.20±0.28                & 93.96±2.62                 & 93.31±2.71              & 99.37±0.22                                                                   & 93.46±1.35                                                                \\
DenseNet169     & 9                                                                      & 14.1M                 & 95.50±0.34                & 94.46±1.81                 & 93.39±2.38              & 99.41±0.15                                                                   & 93.84±1.33                                                                \\
ResNet101       & 8                                                                      & 44.5M                 & 95.23±0.36                & 94.40±1.67                 & 92.77±2.55              & 97.37±4.61                                                                   & 93.49±1.51                                                                \\
VGG16           & 7                                                                      & 138.4M                & 89.70±0.63                & 84.72±3.79                 & 80.93±4.43              & 98.66±0.41                                                                   & 82.25±2.46                                                                \\
ViT-Base        & 6                                                                      & 86.6M                 & 85.40±0.76                & 80.61±5.51                 & 76.52±6.27              & 98.08±0.53                                                                   & 77.51±4.17                                                                \\
DenseNet161     & 5                                                                      & 28.7M                 & 95.50±0.33                & 94.27±2.04                 & 93.82±2.35              & 99.41±0.14                                                                   & 93.98±1.25                                                                \\
VGG19           & 5                                                                      & 143.7M                & 88.92±1.24                & 83.28±5.29                 & 79.03±6.41              & 98.55±0.44                                                                   & 80.34±4.16                                                                \\ \bottomrule
\end{tabular}}
\end{table}

The comprehensive analysis of the metrics demonstrates that Best-EarNet performs well in the task of diagnosing ear lesions. Additionally, its unique network architecture design enables it to achieve ultra-fast average FPS and a remarkably small model parameter size. As anticipated, Best-EarNet effectively balances diagnosis capability, inference speed, and model parameter size.
\begin{figure}[ht]
    \centering
    \includegraphics[width=0.99\linewidth]{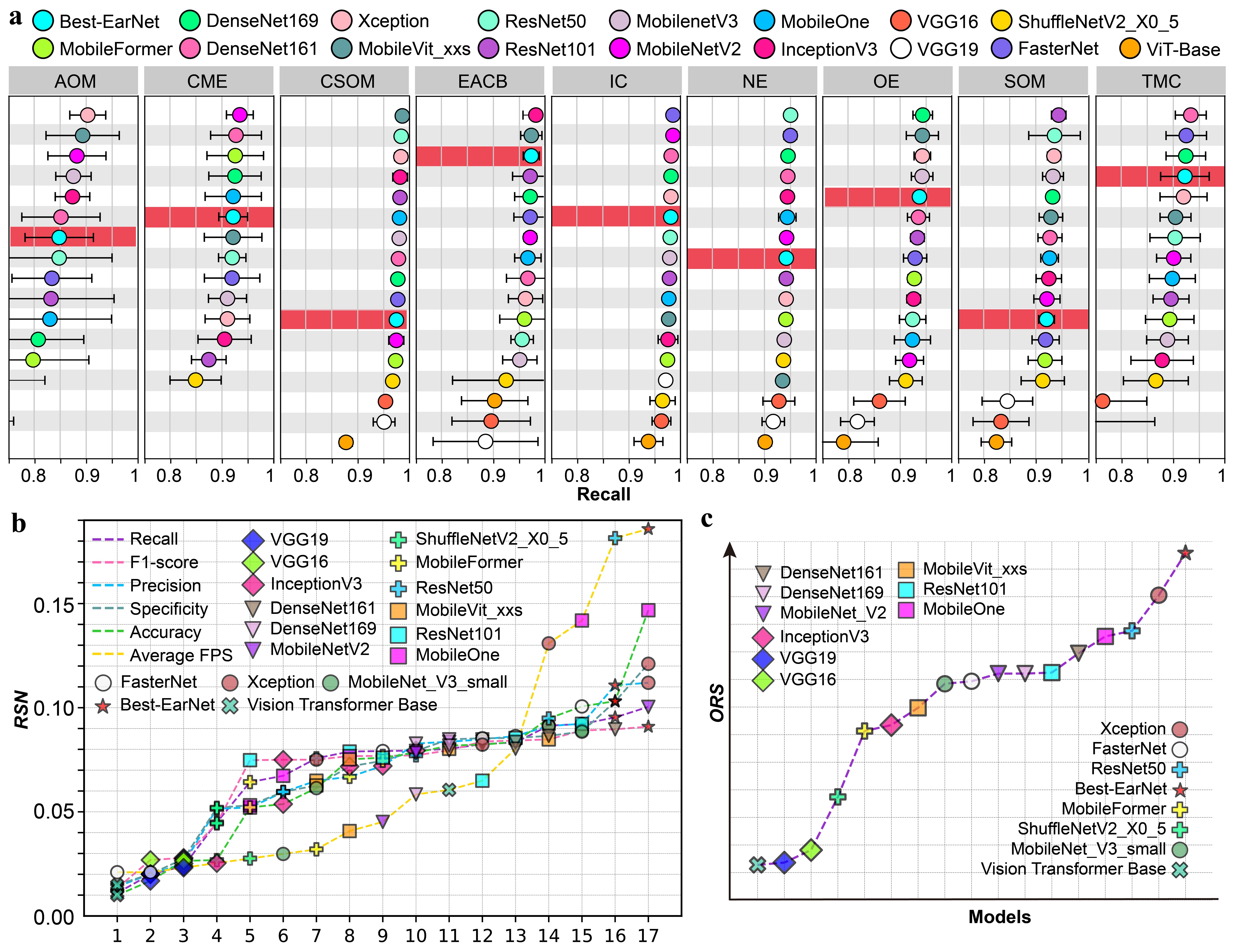}
    \caption{(a) 95\% IC Recall for all models in each category. (b) The RSN of Recall, F1-score, Precision, Accuracy and Average FPS for all models. (c) The ORS of all models.}
    \label{fig5}
\end{figure}
\subsection{Testing on Different Populations and FSH DataSet}
The incidence and severity of ear diseases are often influenced by age and gender. So, to further understand the diagnosis capabilities of the model, detailed testing was conducted using Best-EarNet on various populations categorized by gender and age (Fig. 6a-b). The experimental results demonstrate that Best-EarNet exhibits excellent diagnosis capabilities across all genders and age groups. It is worth noting that due to the relatively small number of images for AOM, Best-EarNet's diagnosis performance for AOM may be slightly lower compared to other conditions.
\begin{figure}[ht]
    \centering
    \includegraphics[width=0.99\linewidth]{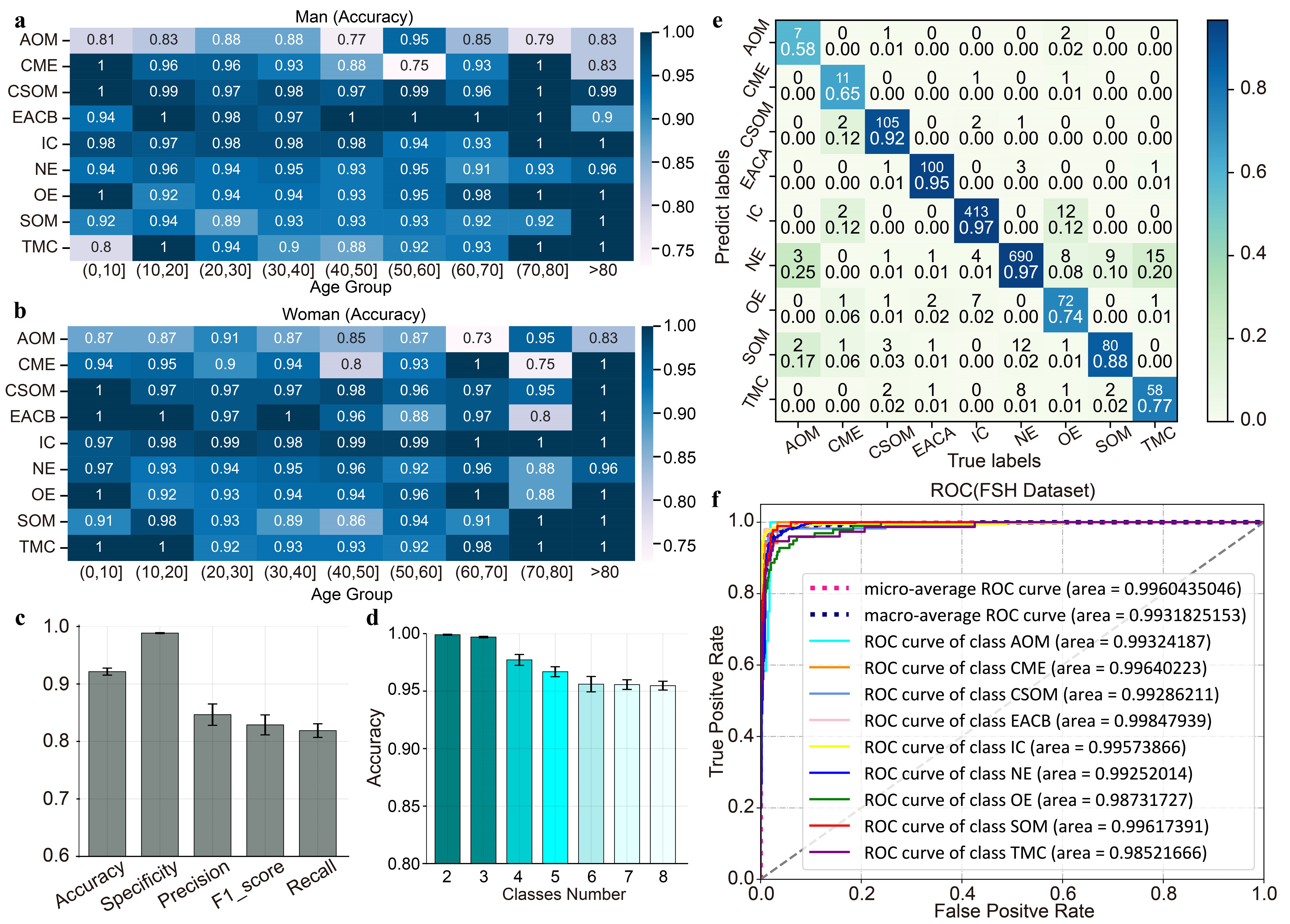}
    \caption{ (a,b) The distribution of testing accuracy of the Best-EarNet in different genders and age groups.(c) Best-EarNet testing in the FSH dataset.(d) Classes extension testing of Best-EarNet. (e) The confusion matrix of FSH dataset calculated by Best-EarNet with 92.98\% Accuracy. (f) The receiver operating characteristic of FSH dataset calculated by Best-EarNet with 92.98\% Accuracy.}
    \label{fig6}
\end{figure}
Furthermore, Best-EarNet was subjected to cross-site testing using the FSH Dataset (Fig. 6c). The five-fold cross-validation yielded accuracy rates of 92.98\%, 91.34\%, 92.43\%, 92.07\%, and 91.89\%, demonstrating high and stable accuracy. As our model's corresponding categories do not cover all types of ear diseases, we conducted category expansion experiments to test the model's applicability for future category expansion. For example, in the N-class testing, we randomly selected N classes from the 9 available classes for model training and testing (each N-class testing needs to be repeated six times). The results were shown in Fig. 6d. The Accuracy for 2 and 3 categories exceeded 99\%, and as the number of categories increased, the accuracy slightly decreased. However, when the number of categories reaches 6 or more, the Accuracy stabilized, remaining above 95\%. This indicates that Best-EarNet has good category expansion capabilities. We selected the model with an accuracy of 92.98\% as the final deployment model. To provide comprehensive data support for our practical deployment, we generated the corresponding confusion matrix (Fig. 6e) and ROC (Fig. 6f).
\subsection{Visual understanding of Best-EarNet decisions}
In the field of medical image processing, model interpretability is crucial for assisted diagnosis, compliance requirements, patient safety, trust building, and model optimization. By explaining the model's prediction results and inference process, the credibility and reliability of medical image processing models can be improved, providing better decision support for healthcare professionals and patients\cite{45}.

In this study, we employed the Gradient-weighted Class Activation Mapping (Grad-CAM) method to visualize and interpret the decision results of Best-EarNet. Grad-CAM is an interpretability technique used to visualize the attention regions within deep neural networks to understand the image regions that the model focuses on during the prediction process. It maps the network's activations to the input image space, generating a heat map. In this process, it calculates the gradients of each output class by backpropagation and multiplies the gradients with the activation feature maps.
\begin{figure}[ht]
    \centering
    \includegraphics[width=0.99\linewidth]{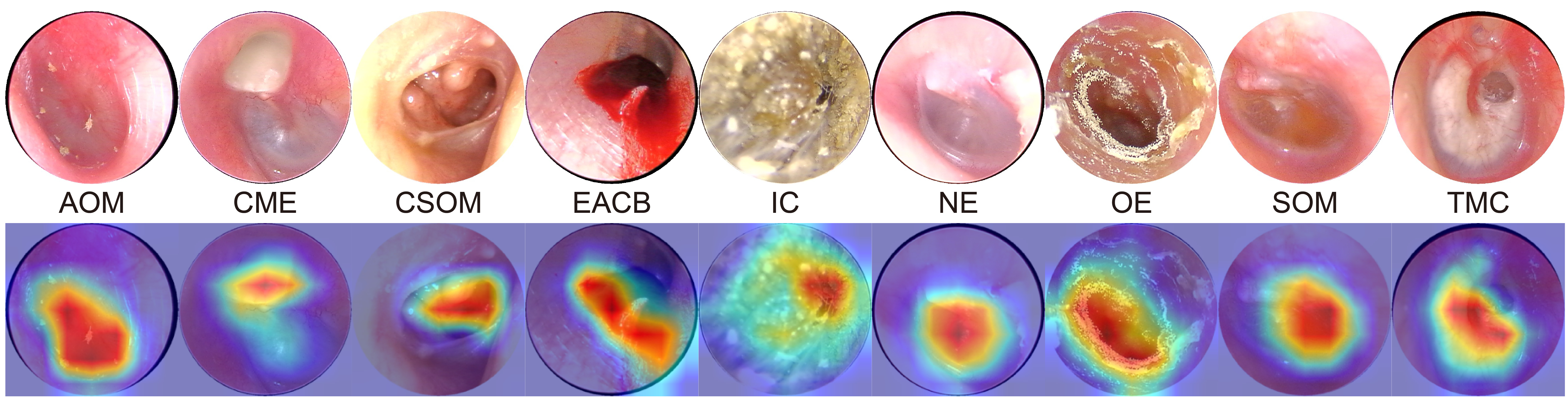}
    \caption{Grad-CAM results of various samples.}
    \label{fig7}
\end{figure}

Fig. 7 presented the results of Grad-CAM. It can be observed that Best-EarNet effectively focuses on the important regions of each class image, consistent with the areas of interest for clinical doctors. These results not only explain the decision process of Best-EarNet but also serve as a reminder for users to pay attention to specific regions in subsequent real-world applications.
\subsection{Comparison of various versions of Ear-Keeper}
After comprehensive model evaluation, we developed four versions of the application using the Java programming language. The Pad and PC versions run on Ubuntu 20.04 and Windows 10 operating systems, respectively. The Mobile (Cloud) and Mobile (Local) versions are designed for the Android platform. The physical products of each version and the operation page as shown in Fig. 8 and Fig. S4, respectively.
\begin{figure}[ht]
    \centering
    \includegraphics[width=0.99\linewidth]{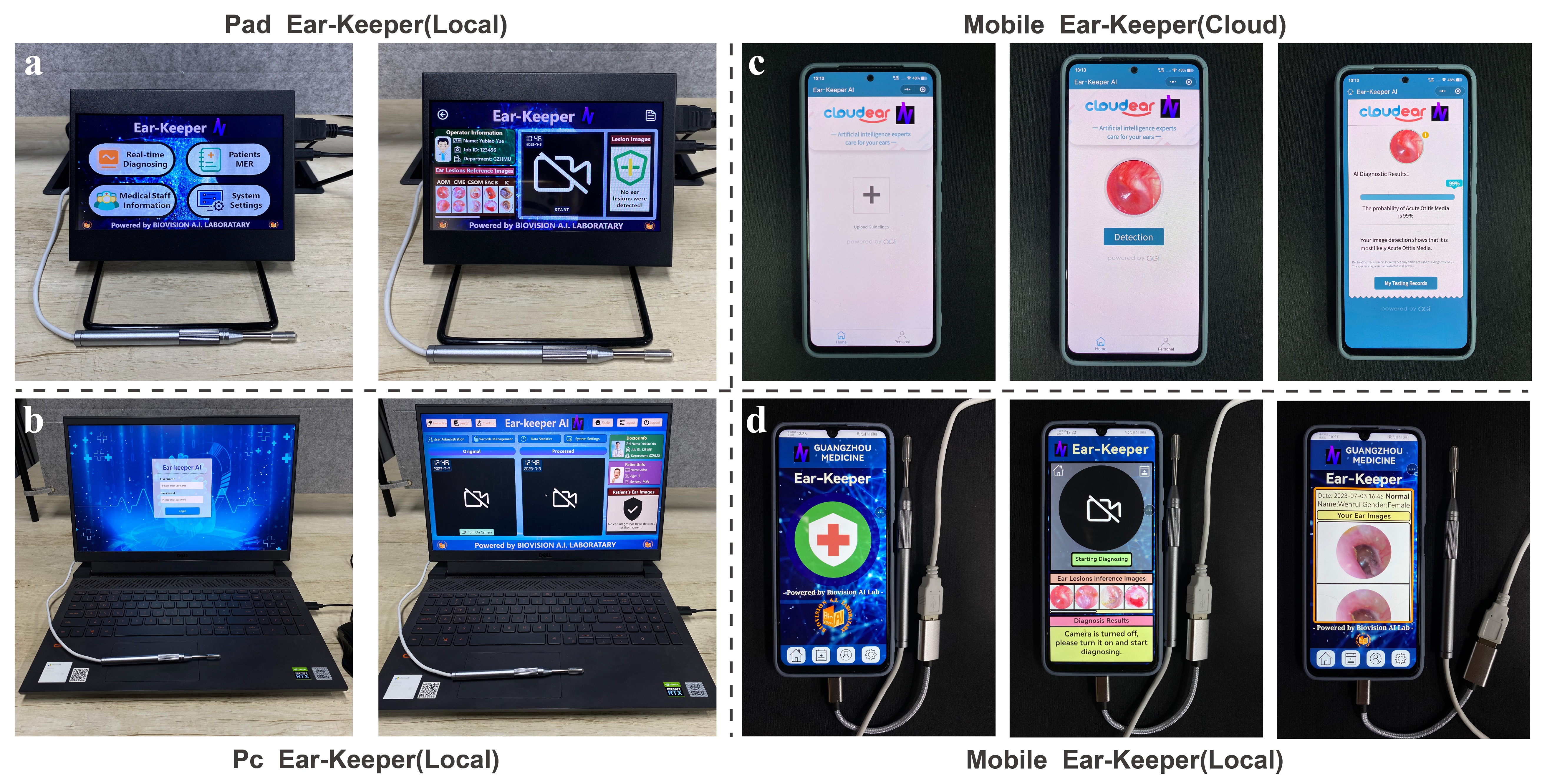}
    \caption{The Application pages of various versions. (a) The main pages of Pad Ear-Keeper(Local). (b) The main pages of PC Ear-Keeper(Local). (c) The main pages of Mobile Ear-Keeper(Cloud). (d) The main pages of Mobile Ear-Keeper (Local).}
    \label{fig8}
\end{figure}
The Pad Ear-Keeper (Local) (Fig. 8(a)) is an integrated device equipped with a 12th Gen Intel(R) Core(TM) i5-1240P CPU. During practical usage, the diagnosis speed ranges from 45 to 56 frames per second (FPS). This device operates independently without requiring an internet connection, making it suitable for medical settings with limited resources. Medical professionals can simply insert the electronic otoscope into the pad to perform real-time monitoring and diagnosis of the ear. The PC Ear-Keeper (Local) (Fig. 8(b)) is an offline application that can be deployed on computers in specialized otolaryngology departments. It contributes to improved efficiency in traditional outpatient examinations and enhances the patient experience. The application was tested on a laptop equipped with an AMD Ryzen 5 4500U CPU with Radeon Graphics. Due to the activation of the heat map display feature, the actual FPS during usage ranges approximately from 25 to 30 (without the heat map feature, the FPS is around 50 to 55). For users who prefer a simpler mobile application, we offer the Mobile Ear-Keeper (Cloud) version (Fig. 8(c)). This version requires an internet connection and involves capturing ear images in advance using the otoscope, which are then uploaded to our server to obtain the corresponding diagnosis results. The Mobile Ear-Keeper (Local) (Fig. 8(d)) is a powerful mobile application similar to the Pad version but more convenient, catering to non-professionals. Users only need to insert the electronic otoscope into their mobile devices to perform real-time ear monitoring. The application retains a record of each diagnosis. We tested this application on a Huawei DVC-AN20 device, which utilizes the MediaTek MT6873 5G CPU. The real-time FPS during usage ranges from 35 to 45. Additionally, when saving images of ear lesions, we employed OpenCV to evaluate image clarity in the PC, Pad, and Mobile (Local) versions. Only images judged as clear were recorded.
\section{Conclusion}
Otolaryngology is a special department in hospitals that deals with a diverse patient population distributed across various departments. However, there exists a significant gap between this department and others in diagnosis. Especially, in the case of ear diseases, other departments may struggle to differentiate between various subtypes of otitis media, leading to misdiagnosis and delayed treatment, such as in pediatrics and emergency medicine. On the other hand, the increasing number of healthcare visits related to ear diseases has resulted in reduced access to diagnosis and treatment opportunities in certain regions, especially in the post-pandemic era. For common ear diseases with high incidence rates, intelligent and convenient diagnosis devices can alleviate these issues and provide more effective diagnoses. They can also enable people in underserved areas to access convenient and reliable examinations.

Since the ear endoscope used for ear examinations can easily be replaced by commonly available electronic otoscopes, which are user-friendly and require no specialized training, emerging artificial intelligence technologies can be readily developed into intelligent ear disease recognition applications. Indeed, there have been studies on artificial intelligence-based ear disease recognition. However, most of the previous research focused on model validation without considering the model's response to specific application scenarios and corresponding devices. Based on this, our study comprehensively evaluated previous research models and developed a model suitable for various scenarios and corresponding devices in the field of ear diseases, including cloud-based applications, PC-based applications, mobile-based applications, Pad-based applications, and other portable device-based applications.

The Ear-Keeper model designed in this study achieved an average Frames Per Second (FPS) of 80M and a model parameter size of 0.77M. Through five-fold cross-validation using 24,432 ear endoscopic images, the model achieved an accuracy of 95.23\%. The model demonstrated excellent performance in tests conducted on different age groups and across different regions. Additionally, based on this model, we developed application systems for various scenarios using different devices, including cloud-based, PC-based, mobile-based, and Pad-based systems. These application systems exhibited good responsiveness in testing. The model developed in this study demonstrated good applicability across various ear disease application scenarios and devices. Furthermore, the model exhibited stable performance in category expansion experiments, indicating its potential for sustainable category expansion.

In summary, our design of the Best-EarNet model strikes a good balance between inference speed, model parameter size, and diagnosis performance. It elevates the application of artificial intelligence in intelligent ear disease recognition to new heights. Moreover, the Ear-Keeper application developed based on Best-EarNet is suitable for various application scenarios, providing a convenient intelligent assistant for a large population in need. Especially for developing countries and regions with a high prevalence of ear diseases, it has the potential to help mitigate the extensive physical, health, and economic losses caused by such conditions.
\section*{Data availability}
The data are not available for public access because of patient privacy concerns but are available from the corresponding author on reasonable request.
\bibliographystyle{unsrt}  
\bibliography{references}  

\end{document}